\documentclass[letterpaper, 10 pt, journal, twoside]{IEEEtran}
\IEEEoverridecommandlockouts

\usepackage{times}
\usepackage{array}
\usepackage{comment}
\usepackage{csquotes}

\usepackage[numbers,sort&compress]{natbib}
\usepackage{multicol}
\usepackage[bookmarks=true]{hyperref}

\usepackage{amsmath}
\usepackage{amssymb}
\usepackage{cancel}
\usepackage{wasysym}
\usepackage{xfrac}
\usepackage{mathrsfs}
\usepackage{euscript}
\usepackage{bbm}

\DeclareMathAlphabet{\mymathbb}{U}{bbold}{m}{n}

\usepackage{amsthm}
\newtheorem{rmk}{Remark}
\newtheorem{thm}{Theorem}
\newtheorem{assumption}{Assumption}
\newenvironment{shortPrf}
  {\noindent \textit{Proof.\space}}
  {\hfill $\square$}

\usepackage{xcolor}
\usepackage{balance}
\usepackage[normalem]{ulem}

\usepackage[font=footnotesize]{caption}
\usepackage{graphicx}
\usepackage{caption}
\usepackage{subcaption}
\usepackage{epstopdf}
\usepackage{float}
\usepackage{tikz}
\usepackage{wrapfig}
\newlength{\sepwid}
\usepackage[maxfloats=256]{morefloats}
\usepackage{url}

\usepackage{lipsum}
\usepackage{cuted}

\newcommand{\real}{\mathbb{R}}
\newcommand{\transp}{\top}

\begin{document}
\bstctlcite{IEEEexample:BSTcontrol}

\title{Self-Healing Distributed Swarm Formation Control Using Image Moments}

\author{
C. Lin Liu$^{1,2}$, Israel L. Donato Ridgley$^{2,3}$, Matthew L. Elwin$^{1,2}$, Michael Rubenstein$^{1,2,4}$, Randy A. Freeman$^{2,3,5}$, and Kevin M. Lynch$^{1,2,5}$

\thanks{This work was supported by NSF CMMI-2024774.}
\thanks{All authors are affiliated with Northwestern University, Evanston, IL 60208 USA. (emails: {\footnotesize lin.liu@u.northwestern.edu}, {\footnotesize freeman@northwestern.edu}, {\footnotesize israelridgley2023@u.northwestern.edu}, {\footnotesize elwin@northwestern.edu}, {\footnotesize rubenstein@northwestern.edu}, {\footnotesize kmlynch@northwestern.edu})

$^1$Department of Mechanical Engineering, $^2$Center for Robotics and Biosystems, $^3$Department of Electrical \& Computer Engineering, $^4$Department of Computer Science, $^5$Northwestern Institute on Complex Systems

We thank Billie Strong and Marko Vejnovic  for their contributions to the Coachbot hardware.}  
} 

\maketitle

\begin{abstract} 
Human-swarm interaction is facilitated by a low-dimensional encoding of the swarm formation, independent of the (possibly large) number of robots. We propose using image moments to encode two-dimensional formations of robots. Each robot knows its pose and the desired formation moments, and simultaneously estimates the current moments of the entire swarm while controlling its motion to better achieve the desired group moments. The estimator is a distributed optimization, requiring no centralized processing, and self-healing, meaning that the process is robust to initialization errors, packet drops, and robots being added to or removed from the swarm. Our experimental results with a swarm of 50 robots, suffering nearly 50\% packet loss, show that distributed estimation and control of image moments effectively achieves desired swarm formations.
\end{abstract}


\section{Introduction} 

Swarms of hundreds or thousands of robots, each implementing simple behaviors, can generate complex and useful group behaviors~\cite{Rubenstein2014,IOC2021}.
In applications such as environmental monitoring and search and rescue, humans may need to control the swarm formation in real time~\cite{Schranz2020}. To facilitate this one-to-many interaction, the interface should impose a low cognitive burden, and the human should not need to directly specify each robot's motion~\cite{Kolling2016}. In other words, the intended formation should be representable using a small number of variables, independent of the number of robots.

\begin{figure} 
\centering
$\vcenter{\hbox{\fbox{\subfloat{\includegraphics[width=0.08\textwidth]{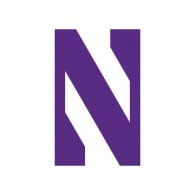}}}}}$
$\vcenter{\hbox{\subfloat{\includegraphics[width=0.38\textwidth]{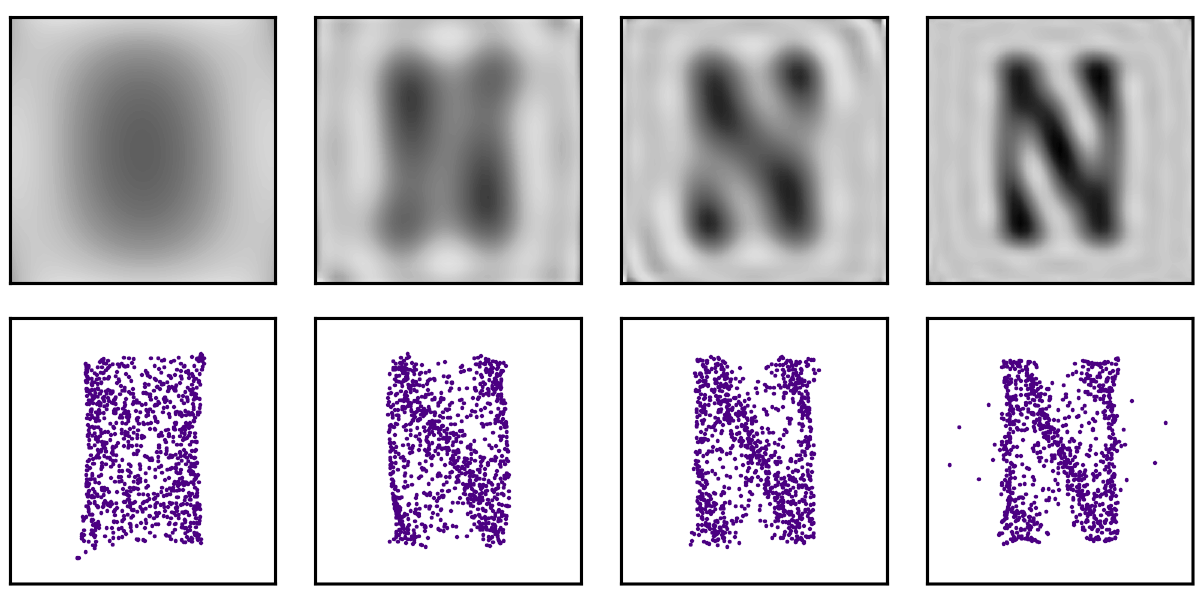}}}}$
\caption{The left image specifies a desired two-dimensional swarm distribution. White indicates zero robot density and purple indicates a constant positive density. Top row: Standard computer vision image reconstruction using representations of the desired distribution based on Legendre moments of up to fifth, tenth, 15th, and 20th order, respectively. Bottom row: The final configurations of 1000 simulated robots after moving so their collective Legendre moments, up to fifth, tenth, 15th, and 20th order, respectively, approximately match the desired Legendre moments.}
\label{fig:N}
\end{figure}

In this paper, we introduce low-dimensional representations of swarm formations based on finite sets of moments of a density distribution. Such moments can describe the center of mass location of a distribution (first-order moments), the principal axes of the distribution (second-order moments), and increasingly fine, high-frequency features of the distribution as the moment order increases. These distributions can be defined on spaces of arbitrary dimensions; for example, a drone formation can be described as a distribution over the six-dimensional space of rigid-body configurations. In this paper, we focus on formations of robots modeled as points in a plane. This allows us to take advantage of \emph{image moment} representations of planar distributions (Section~\ref{sec:moments}), an image compression technique pioneered in computer vision (Figure~\ref{fig:N}).

In our setup, image moments, representing a desired formation, are transmitted to the robots via broadcast. Each robot continuously senses its position and communicates with neighbors to estimate the moments of the entire swarm. Using these estimated moments, the robot locally calculates a motion control that drives the swarm to achieve the desired moments, and therefore the desired formation. 
We require the robots' estimation and control to be (a) \emph{distributed}, meaning that there is no centralized controller that can be a single point of failure; (b) \emph{scalable}, meaning that the length of the messages between the robots is constant (independent of the number of robots); and (c) \emph{self-healing}~\cite{Ridgley2021}, which imparts robustness to packet loss and transient computation errors, 
and allows robots to be added to or subtracted from the swarm at any time.

\subsection{Related Work} 

\subsubsection{Human-Swarm Interaction} 
A human can influence swarm shape through neighbor interactions by controlling the motion of one robot~\cite{Bashyal2008, Brown2014} or a subgroup of robots using beacons that broadcast desired local behaviors such as attracting and repulsing motions~\cite{Kolling2013}. These methods are not scalable since the human's influence decreases as the swarm size increases. Kira and Potter describe a bottom-up control scheme with virtual agents to influence the swarm through local interactions, and a top-down control scheme that changes global swarm characteristics to influence individual robot behaviors~\cite{Kira2009}. 
Egerstedt et al.\ describe one approach where a human controls leader robots to change the configuration of a leader-follower swarm, and a second approach where swarm robots are treated as particles suspended in a fluid comprised of regions that a human can influence~\cite{Egerstedt2014}. 

A human can specify regions of interest for the robots rather than individual robot locations~\cite{Saska2016}. Prabhakar et al.\ specified regions of interest by shading map areas on a tactile tablet for robots to optimally cover using ergodic control~\cite{Prabhakar2020}. A human can also specify time-varying density distributions~\cite{Lee2015}.

\subsubsection{Image Moments} 
Image moments encode a density distribution over a plane, $f(x,y): \real \times \real \rightarrow \real$, using the weighted sum of low-degree basis polynomials on the plane~\cite{Pawlak2006, Qi2021}. The weights are called \emph{moments}. Example orthogonal basis polynomials include Legendre polynomials, Zernike polynomials, and pseudo-Zernike polynomials. The higher the degree of the polynomials included in the representation, the greater the detail of the encoded distribution~\cite{Teh1988}.  

In general, pseudo-Zernike moments (PZMs) represent greater detail than Zernike moments (ZMs) of the same order~\cite{Hosny2014, Mukundan1998, Hosaini2013}. The performance of PZMs vs.\ Legendre moments (LMs) depends on the image. When reconstructing the letter ``E,'' PZMs have lower reconstruction error and higher robustness to image noise~\cite{Teh1988}. For an eye iris image, LMs are more robust to white Gaussian noise~\cite{Hosaini2013}.

\subsubsection{Low-Dimensional Representations of Formations and Distributed Formation Control} 
A shape can be defined by goal locations that are assigned to each robot in a swarm~\cite{Wang2020}. This approach is not scalable since each robot needs a defined position. 
A different, more scalable approach is to use attractive and repulsive potential functions to create desired formations~\cite{Jung2014, Yang2022}. 
Fu et al.\ use a consensus algorithm to keep robots in a consistent distribution relative to a virtual leader that moves along a given trajectory~\cite{Fu2020}.

In previous work, a decentralized swarm of eight robots was used to achieve first- and second-order inertial moments with convergence guarantees~\cite{Yang2008, Hwang2013}. The robots were also able to achieve a series of goal formations to move past an obstacle. 

\subsection{Contributions} 
We present simultaneous distributed estimation and control of a swarm formation using LMs and PZMs. The approach is scalable and self-healing in theory, simulation, and experiments with up to 50~robots. Compared to previous work (e.g.,~\cite{Yang2008, Hwang2013}), the proposed approach gives moments higher than second-order and therefore greater detail in specifying the formation; provides robustness to packet loss; capitalizes on a recent method for distributed optimization that reduces the amount of robot-robot communication needed for estimation~\cite{Ridgley2021}; and has been verified on large experimental swarms.

\section{Distributions Using Image Moments}
\label{sec:moments}

\subsection{Legendre Moments (LMs)}
The Legendre polynomials~\cite{Teh1988,Teague1980,Mukundan1995} 
$P_m(x)$ of order $m = 0, 1, ..., \infty$ are defined on the domain $[-1,1]$ recursively using Bonnet's recursion formula~\cite{Mukundan1995},
\begin{equation}
\begin{gathered}
    P_0(x) = 1, \; P_1(x) = x, \\
    P_m(x) = \frac{2m-1}{m} x P_{m-1}(x) - \frac{m-1}{m}P_{m-2}(x), \; m\geq 2.
\end{gathered}
\end{equation}
With these polynomials, the orthogonal basis functions of order $n = p+q$ over the square $(x,y) \in [-1,1] \times [-1,1]$ have the form $P_p(x) P_q(y)$ where $p,q = 0, 1, \ldots, \infty$. Any distribution $f(x,y)$ can be written as a linear combination of these basis functions
\begin{equation}
    f(x,y) = \sum_{p=0}^\infty \sum_{q=0}^\infty M_{pq} P_p(x) P_q(y),
\end{equation}
where the \emph{Legendre moments} (LMs) $M_{pq}$ for a distribution $f(x,y)$ are calculated as
\begin{equation}
    M_{pq} = \frac{(2p+1)(2q+1)}{4} \int_{-1}^{1} \int_{-1}^{1} P_p(x)P_q(y)f(x,y) \,dx \,dy.
    \label{eq:lpoly_integral}
\end{equation}

For a discrete distribution represented by point masses at the centers of $(R \times C)$ ``pixels,'' the LMs can be written
\begin{equation}
    M_{pq} = \frac{(2p+1)(2q+1)}{4}\sum_{i=1}^{C} \sum_{j=1}^{R} P_p(x_i) P_q(y_j) \mu_{ij},
\end{equation}
where $(x_i,y_j)$ is the location of pixel $(i,j)$ and $\mu_{ij}$ represents the mass at pixel $(i,j)$. Similarly, for a set of $N$ point robots of unit mass at locations $(x_i,y_i), i = 1, \ldots, N$, the LMs can be written as
\begin{equation}
    M_{pq} = \frac{(2p+1)(2q+1)}{4} \sum_{i=1}^N P_p(x_i) P_q(y_i).
    \label{eq:Lmoments}
\end{equation}
A distribution $f(x,y)$ can be approximately reconstructed using its LMs truncated at order $n$,
\begin{equation}
    f(x,y) \approx \sum_{p = 1}^{n} \sum_{q=0}^p M_{p-q,q} P_{p-q}(x) P_q(y),
\end{equation}
where the fidelity of the reconstruction improves as $n$ increases. 
The single zeroth-order moment is the total mass of the distribution, the two first-order moments correspond to the location of the center of mass, and the three second-order moments relate to the principal axes of the distribution. We do not use zeroth-order moments in our representations to allow them to be independent of the number of robots. At each order $n$, there are $n+1$ LMs, so a formation represented by all moments from first to $n$th has $\mathfrak{m}=n(n+3)/2$ elements.

\subsection{Pseudo-Zernike Moments (PZMs)}

Zernike polynomials form an orthogonal basis for functions defined on the unit disk $x^2+y^2 \leq 1$~\cite{Teh1988}. \emph{Pseudo-Zernike polynomials} form a related orthogonal basis on the unit disk while providing improved image reconstruction properties~\cite{Teh1988}. A pseudo-Zernike polynomial $W_{pq}$ is defined as~\cite{Hosny2014}
\begin{equation}
    W_{pq}(r,\theta) = S_{pq}(r)e^{\mathfrak{j}q\theta},
\end{equation}
where $p = 0, 1, \ldots, \infty$ is the order of the radial polynomial, $q = 0, 1, \ldots, p$ is the degree of angular repetition, $r \leq 1$ is the radius, $\theta$ is the angle, and $\mathfrak{j}$ is the imaginary unit. The radial function $S_{pq}(r)$ is
\begin{equation}
    S_{pq}(r) = \sum_{k=q}^{p} B_{pqk} r^k,
\end{equation}
where the coefficients $B_{pqk}$ are 
\begin{equation}
    B_{pqk} = \frac{(-1)^{(p-k)}(p+k+1)!}{(p-k)!(q+k+1)!(k-q)!}.
\end{equation}

Any distribution $f(r,\theta)$,\; $r \leq 1$, can be expressed as a linear combination of the pseudo-Zernike basis functions $W_{pq}(r,\theta)$
\begin{equation}
    f(r,\theta) = \sum_{p=0}^\infty \sum_{q=0}^{p} M_{pq} W_{pq}(r,\theta),
\end{equation}
where the \emph{pseudo-Zernike moments} (PZMs) $M_{pq}$ are
\begin{equation}
    M_{pq} = \frac{p+1}{\pi} \int_{0}^{2\pi} \int_{0}^{1} W^*_{pq}(r,\theta) f(r,\theta) \, r\,dr \,d\theta,
\end{equation}
and $^*$ indicates the complex conjugate~\cite{Teh1988}.

Analogously, the PZMs are 
\begin{equation}
    M_{pq} = \frac{p+1}{L_p}\sum_{i=1}^{C} \sum_{j=1}^{R} W^*_{pq}(r_{ij},\theta_{ij}) \mu_{ij}
\end{equation}
for an $(R \times C)$-pixel distribution, where $r_{ij} = (x_{ij}^2 + y_{ij}^2)^{\frac{1}{2}}$, $\theta_{ij} = \operatorname{atan2}(y_{ij},x_{ij})$, the mass $\mu_{ij}$ is zero for $r_{ij}>1$, and 
$L_p$ is the total number of pixels satisfying $r_{ij} \leq 1$. For $N$ unit-mass robots, the PZMs are 
\begin{equation}
    M_{pq} = \frac{p+1}{L_p} \sum_{i=1}^N W^*_{pq}(r_i,\theta_i).
    \label{eq:PZmoments}
\end{equation}
A distribution $f(r,\theta)$ can be approximately reconstructed using its PZMs truncated at radial order $n$,
\begin{equation}
    f(r,\theta) \approx \sum_{p=1}^n \sum_{q=0}^p M_{pq} W_{pq}(r,\theta),
\end{equation}
where the fidelity of the reconstruction improves as $n$ increases. Just like the LMs, there are $n+1$ PZMs at radial order $n$, and the total number of PZMs from first- to $n$th-order is $\mathfrak{m}=n(n+3)/2$.

\section{Estimation and Control}
The position of the $i$th point robot in a swarm of $N$ robots is $s_i=[s_{ix}, s_{iy}]^\transp \in \real^2$, and the swarm configuration is the column vector $s=[s_1^\transp,s_2^\transp,\ldots,s_N^\transp]^\transp \in \real^{2N}$. Each robot senses its own position and knows the desired formation expressed as a finite set of image moments. All robots estimate the current swarm moments based only on local measurements and messages from neighbors. The size of these messages depends on the number of image moments used in the representation, but not the number of robots. This is important for scaling up to formations of thousands or millions of robots. Robots calculate their own motions based on their estimated moments. 

The error between distributions can be expressed as:
\begin{itemize}
\item \textbf{Moment vector error:} Given $\mathfrak{m}$-vector moments for two formations, $M_1=[M_{11},\ldots, M_{1\mathfrak{m}}]^\transp$ and $M_2=[M_{21},\ldots, M_{2\mathfrak{m}}]^\transp$, this error is defined $M_1-M_2 \in \real^\mathfrak{m}$.

\item \textbf{Mean-square reconstruction error (MSRE):} The MSRE is defined using pixel-based image reconstructions from the moment vectors $M_1$ and $M_2$ as
\begin{equation}
    \textrm{MSRE} = \frac{\sum_i \sum_j (f_{M_1}(x_i,y_i) - f_{M_2}(x_i,y_i))^2}{\sum_i \sum_j f^2_{M_2}(x_i,y_j)},
\end{equation}
where $M_2$ corresponds to the desired moments. To calculate this error, we evaluate the two reconstructions at $41 \times 41$ points corresponding to increments of $0.05$ in each dimension on the domain $(x,y) \in [-1,1] \times [-1,1]$. For PZMs, points outside the unit circle are not used. 
\end{itemize}

\begin{assumption}
    We assume $\mathfrak{m} < 2N$ so a moment vector $M \in \real^{\mathfrak{m}}$ is a compressed representation of a distribution.
\end{assumption}

\begin{rmk}
Note that the distribution reconstruction for a given moment vector $M$ is unique, but since $M$ is a compressed representation of a formation, in general there are infinitely many robot formations that have the same moments.
\end{rmk}

\begin{figure} 
\centering
\includegraphics{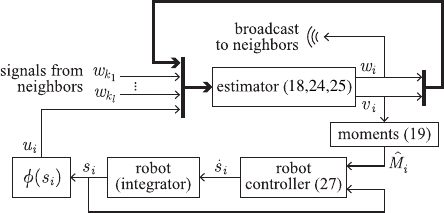}
\caption{Block diagram for robot $i$.}
\label{fig:block_diagram}
\end{figure}

\subsection{Distributed Moment Estimation} 
The network of $N$ robots in a swarm is represented as a directed graph consisting of edges $E$ and vertices $V={1,\ldots,N}$. An edge $E_{ij}$ exists between any two robots $i$ and $j$ when $i$ can send messages to $j$. 
Each robot $i$ knows the number of robots it sends a message to, i.e., its \emph{out-degree} $d_i^{\textrm{out}}$. The out-Laplacian is $\mathfrak{L}=D^{\textrm{out}}-A_\mathrm{adj}$, where the out-degree matrix $D^\textrm{out}$ is the diagonal matrix $D_{ii}^\textrm{out}=d_i^\textrm{out}$ for all $i \in V$, and an element $A_{ij}$ of $A_\mathrm{adj}$ is $1$ if $E_{ij}$ exists and zero otherwise. 

Robot $i$'s estimator and controller is illustrated in Figure~\ref{fig:block_diagram}. The estimator maintains two local signals, $w_i \in \real^{\mathfrak{m}+1}$ and $v_i \in \real^{\mathfrak{m}+1}$. The robot broadcasts $w_i$ to its neighbors, receives communications $w_{k_1}, \ldots, w_{k_l}$ from its neighbors $k_1 \ldots k_l$, measures its own position $s_i \in \real^2$, and uses $s_i$ to construct its input vector 
\begin{equation}
u_i(s_i) \equiv \begin{bmatrix}
    \phi(s_i) \\
    1
\end{bmatrix} \in \real^{\mathfrak{m}+1},
\label{eq:estInput}
\end{equation}
where $\phi(s_i) \in \real^{\mathfrak{m}}$ is the \emph{moment-generating function} that generates robot $i$'s contributions to the swarm moment vector, i.e., $\mathfrak{m}$ components of the form $M_{pq}$ from Equation~\eqref{eq:Lmoments} (LMs) or Equation~\eqref{eq:PZmoments} (PZMs) for the single robot $i$.  

For the purpose of analysis, we model estimation and control as synchronous discrete-time processes.
Based on our recent work on self-healing distributed optimization~\cite{Ridgley2020,Ridgley2021}, the robot runs a push-sum algorithm~\cite{Hadjicostis2018} to estimate the swarm's current moment vector. At each time step $t$, $v_i[t]$ and $w_i[t]$ are calculated as
\begin{align}
    v_i[t] &= u_i[t] - d_i^{\textrm{out}} w_i[t] + \sum_{k \in \mathcal{N}_{i}^{\textrm{in}}} w_k[t]
    \label{eq:estimator} \\
    w_i[t+1] &= w_i[t] + \gamma v_i[t], \label{eq:estimator2}
\end{align}
where $\mathcal{N}_i^{\textrm{in}}$ is the set of in-neighbors of robot $i$, and the constant gain $\gamma$, which is the same for all robots, is chosen to satisfy $\gamma d_i^{\textrm{out}} < 1, \forall i$.\footnote{Since $d_i^{\textrm{out}} \leq N-1$, a bound on $N$ suffices to place a bound on $\gamma$.} Robot $i$'s estimate of the swarm's moments is
\begin{equation}
    \hat{M}_i = \left(\frac{1}{v_{i,\mathfrak{m}+1}}\right) v_{i,1\ldots \mathfrak{m}} \in \real^{\mathfrak{m}},
    \label{eq:moments}
\end{equation}
where $v_{i,1\ldots \mathfrak{m}} \in \real^{\mathfrak{m}}$ refers to the first $\mathfrak{m}$ components of $v_i$ and $v_{i,\mathfrak{m}+1} \in \real$ refers to the last component of $v_i$.

The individual robots' signals can be stacked as
\begin{equation}
    u[t] = \begin{bmatrix} u_1[t] \\ \vdots \\ u_N[t] \end{bmatrix}, \;  
    w[t] = \begin{bmatrix} w_1[t] \\ \vdots \\ w_N[t] \end{bmatrix}, \;  
    v[t] = \begin{bmatrix} v_1[t] \\ \vdots \\ v_N[t] \end{bmatrix}
\end{equation}
to write the updates of the whole system as 
\begin{align}
    v[t] &= u[t] - (\mathfrak{L}^\transp \otimes I)w[t] \label{eq:wholesysv} \\
    w[t+1] &= w[t] + \gamma v[t], \label{eq:wholesysw}
\end{align}
where $\mathfrak{L}$ is the out-Laplacian, $\otimes$ is the Knonecker product, and $I$ is the identity matrix.

\begin{thm}
For a strongly-connected constant digraph network, the estimator of \eqref{eq:wholesysv}-\eqref{eq:wholesysw}, 
and a constant input $u[t] = u$, each $v_i$ converges exponentially to 
\begin{equation}
    v_i^* = \begin{bmatrix} z_i \sum_{j=1}^{N} \phi(s_j)\\ z_i N \end{bmatrix},
\end{equation}
where $z_i$ is the $i$th component of the vector $z \in \real^N$ that satisfies $\mathfrak{L}^\transp z=0$ and $\mymathbb{1}^\transp z=1$. 
Therefore $\hat{M}_i$ (Equation~\eqref{eq:moments}) converges to the correct moments $M(s)$.
\label{thm:est-convergence}
\end{thm}

\begin{shortPrf}
    The proof follows from Theorem~1 of~\cite{Ridgley2020}.
\end{shortPrf}

\begin{rmk}
The convergence described in Theorem~\ref{thm:est-convergence} resumes after transients caused by network changes, sensing errors, or packet drops. If the input $u$ is continuously changing, a bound on the convergence error grows with the maximum rate of change of $u$~\cite{Kia2018}. 
\end{rmk}

To address the practical issue of packet drops in a fielded system, we can add memory to the estimator~\cite{Hadjicostis2016}. The memory state $\mathfrak{M}_{ki}$ stores the last message robot $i$ received from neighbor $k$. If no message was received from neighbor $k$ this iteration, its previous message is used in estimation. 
 Equation~\eqref{eq:estimator} can be rewritten as 
\begin{align}
v_i[t] &= u_i[t] - d_i^{\textrm{out}} w_i[t] + \sum_{k \in N_{i}^\textrm{in}} 
\mathfrak{M}_{ki}[t]
\label{eq:memstate} \\ 
\mathfrak{M}_{ki}[t] &= 
    \begin{cases}
        \mathfrak{M}_{ki}[t-1] & \text{if packet lost}\\
        w_k[t] & \text{if packet received}.
    \end{cases}
\end{align}
If a robot has not heard from a neighbor after a set number of consecutive iterations, that neighbor's value in the memory state is removed. This forgetting factor allows the estimator to adjust to changes in the communication network. 

To test the performance of moment estimation, we simulated a network of 50 stationary robots randomly distributed over an arena of size $[-1,1] \times [-1,1]$. The estimators were tested using all-to-all communication networks and networks defined by a communication radius of 0.5, and
for varying levels of packet loss, with and without memory. A packet loss rate of 30\%, for example, indicates that each individual message has a 30\% chance of being dropped. The estimator gain was $\gamma = 1/N$.

\begin{figure}
\centering
\includegraphics[width=\linewidth]{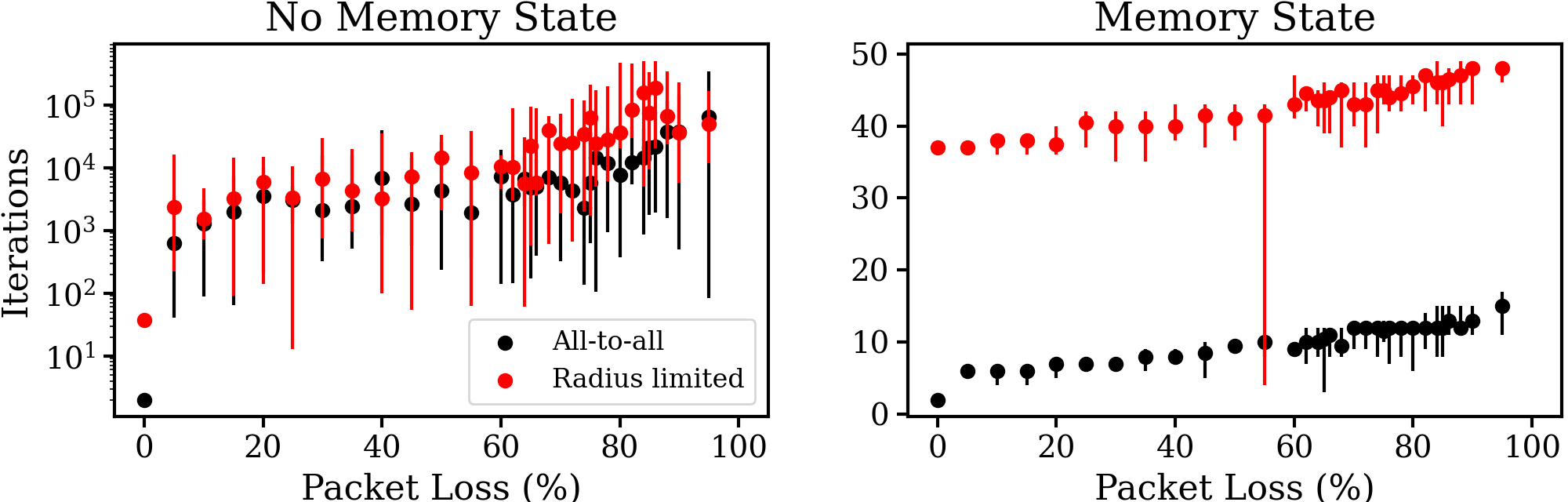}
\caption{Time to convergence for eighth-order LM estimation.}
\label{fig:lePcktLoss}
\end{figure}

\begin{figure}
\centering
\includegraphics[width=\linewidth]{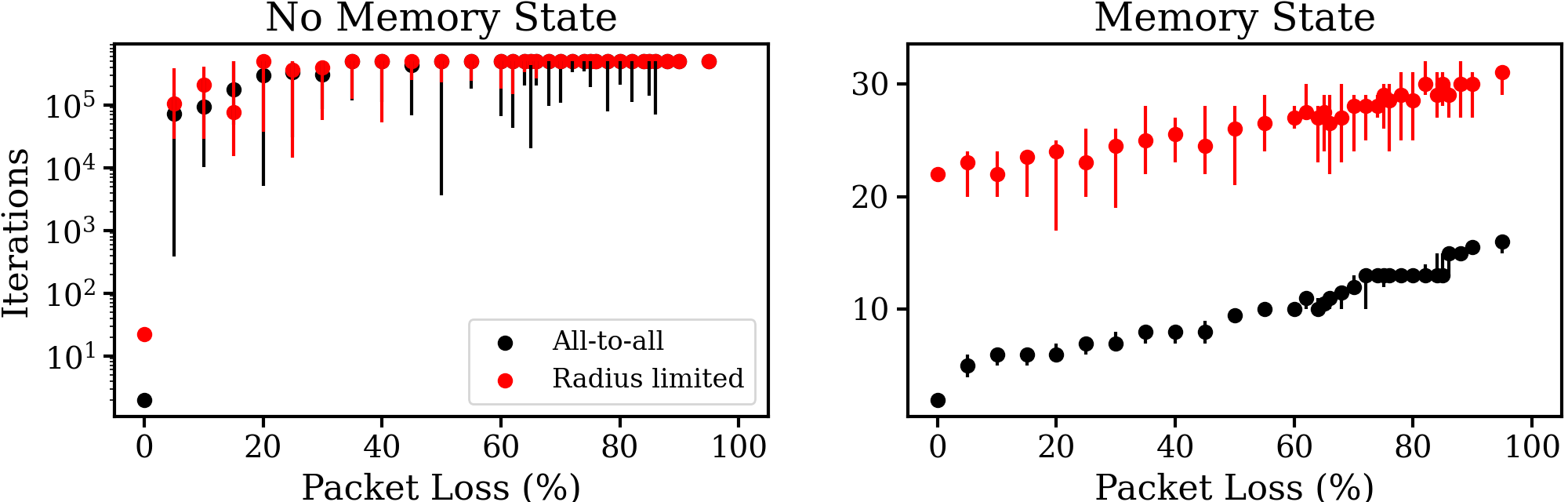}
\caption{Time to convergence for eighth-order PZM estimation.}
\label{fig:pzmPcktLoss}
\end{figure}

Ten trials were conducted for each packet loss level and network type. The robot positions were the same across all trials. The convergence times of estimating LMs and PZMs up to eighth-order are reported in Figures~\ref{fig:lePcktLoss} and \ref{fig:pzmPcktLoss}, respectively. Convergence is considered to be achieved when the error relative to the desired moments is less than $0.01$. If convergence is not reached within 500,000 steps, the trial is terminated. For trials with memory, a neighbor's message is kept in the memory state until $75$ consecutive iterations passed without receiving a message from the neighbor. The all-to-all network results are shown in black while the radius-limited results are shown in red. The point represents the median convergence time of the ten trials and the vertical line indicates the range of convergence times. Memory reduces the convergence time by orders of magnitude when packets are lost and the all-to-all communication network has faster convergence times for both LMs and PZMs.

\subsection{Control of Swarm Moments} 
The control goal is to minimize the error between the estimated and desired moments as given by the quadratic cost
\begin{equation}
C(s)=\frac{1}{2}[M(s)-M^*]^\transp \Gamma [M(s)-M^*],
\label{eq:costfn}
\end{equation}
where $M(s)$ is the vector of moments of the swarm's current configuration $s$, $\Gamma \in \real^{\mathfrak{m} \times \mathfrak{m}}$ is a symmetric positive-definite gain matrix, and $M^*$ is the vector of desired moments. 

Each robot moves according to the gradient control law
\begin{equation}
\dot{s}_i = -[\mathscr{J}(\phi(s_i))]^\transp \Gamma [\hat{M}_i - M^*],
   \label{eq:controllaw}
\end{equation}
where $\hat{M}_i$ is robot $i$'s current estimate of the swarm moments and $\mathscr{J}(\phi)$ is the Jacobian of the moment-generating function $\phi$ with respect to $s_i$. 

To analyze the collective behavior of the control law~\eqref{eq:controllaw}, we consider the case where each robot has a perfect estimate of the current swarm moments, $\hat{M}_i = M(s)$, and we define the moment error $e(s)=M(s)-M^*$. Let $J(s)= \frac{de}{ds} \in \real^{\mathfrak{m}\times 2N}$. The $N$ copies of \eqref{eq:controllaw} can be stacked to give
\begin{equation}
    \dot{s}=-J^\transp (s) \Gamma e(s) \in \real^{2N}. \label{eq:sdynamics}
\end{equation}
Clearly every point $s_e$ for which $e(s_e)=0$ is an equilibrium for \eqref{eq:sdynamics}. We will call these the ``good'' equilibria, in contrast to equilibria where the error is nonzero.  To analyze the local behavior of the system \eqref{eq:sdynamics} near a good equilibrium, we first note that
the time derivative of $e$ is
\begin{equation}
    \frac{de}{dt}=A_e(s)e \;\; \textrm{where} \;\; A_e(s) = -J(s)J^\transp (s) \Gamma.
\label{eq:errordynamics}
\end{equation}

\begin{thm}
Let $s_e$ be a good equilibrium
for the dynamics~\eqref{eq:sdynamics}, and suppose $\operatorname{rank}(J(s_e)) = \mathfrak{m}$. Then $s_e$ is stable in the sense of Lyapunov. Moreover, if the initial configuration $s(0)$ is sufficiently close to $s_e$, then $e$ converges asymptotically to zero.
\label{thm:2}
\end{thm}

\begin{shortPrf}
By assumption $\operatorname{rank}(J(s_e)) = \mathfrak{m}$ and $\Gamma>0$, which together imply that
the matrix $A_e(s_e)$ in \eqref{eq:errordynamics} is Hurwitz.  Therefore there exists a unique positive definite solution $P$ to the Lyapunov equation

\begin{equation}
    PA_e(s_e)+A_e(s_e)^\transp P = -I.
\end{equation}

Let $\widetilde{A}(s)=A_e(s)-A_e(s_e)$; then we can write the derivative of the Lyapunov function
$V_e(e)=e^\transp P e$ along solutions to \eqref{eq:errordynamics} as
\begin{equation}
    \dot{V_e}=-||e||^2 + e^\transp [P \widetilde{A}(s)+\widetilde{A}(s)^\transp P]e.
    \label{eq:vdot12}
\end{equation}
Because $\operatorname{rank}(J(s_e)) = \mathfrak{m}$, there exists a $(2N-\mathfrak{m})\times 2N$ matrix $T$ such that $[T^\transp\;\;J(s_e)^\transp]$ is invertible.  This implies that the mapping from $s$ to $(Ts-Ts_e,e(s))$ is a diffeomorphism in a neighborhood of $s_e$.  Setting $y=Ts-Ts_e$, we can write the system \eqref{eq:sdynamics} in the local $(y,e)$-coordinates as
\begin{align}
    \dot{y}&=C_e(e,y)e \label{eq:edot1}\\
    \dot{e}&=A_e(s_e)e+B_e(e,y)e,
\label{eq:edot}
\end{align}
where $B_e(0,0)=0$.  Because $A_e(s_e)$ is Hurwitz, we can use a Lyapunov function of the form $||y||^2+\sqrt{V(e)}$ to show that the point $(y,e)=(0,0)$ is stable in the sense of Lyapunov for the system \eqref{eq:edot1}--\eqref{eq:edot} (see \cite[Corollary 8.1]{Khalil2002} for details).
Thus the equilibrium $s_e$ for the system \eqref{eq:sdynamics} is stable in the sense of Lyapunov, which means if $s(0)$ is sufficiently close to $s_e$ then $\widetilde{A}(s)$ will remain small in forward time.  Thus from \eqref{eq:vdot12} we obtain $\dot{V}<0$ for $e\neq 0$, and we conclude that $e$ converges asymptotically to zero.
\end{shortPrf} 

To analyze the global behavior of the system \eqref{eq:sdynamics}, we first note that it is a gradient flow.
Hence if $e$ is real analytic and proper, which it is for LMs and PZMs, then every trajectory of this gradient system converges to an equilibrium~\cite{Loj83}. Thus from Theorem~\ref{thm:2} we see that any sufficiently small perturbation away from a good equilibrium will cause the system to converge to a (possibly different) good equilibrium.  Moreover, it follows from \eqref{eq:sdynamics} that 
$\operatorname{rank}(J(s_e)) < \mathfrak{m}$ at any ``bad'' equilibrium $s_e$, i.e., any equilibrium $s_e$ for which $e(s_e)\neq 0$.  Thus the bad equilibria occur only at singular configurations where $J$ loses rank.  Nevertheless, it may be possible for a bad equilibrium to be a local minimum of the cost in \eqref{eq:costfn} and thus be a stable equilibrium for the flow \eqref{eq:sdynamics}.  While we have not found this to be an issue in practice, the problem of how to choose the gain matrix $\Gamma$ to make bad equilibria unstable, or at a minimum to shrink their regions of attraction, is a topic for future research.

The controller was tested by simulating a network of 50 robots in an arena of size $[-1,1] \times [-1,1]$. Each robot had access to the correct swarm moments, $\hat{M}_i=M(s)$, and its motion was simulated using a discretized version of the control law~\eqref{eq:controllaw}:
\begin{equation}
    s_i[t+1] = s_i[t] + \dot{s}_i.
\end{equation}

\begin{figure}
\centering
\includegraphics[width=\linewidth]{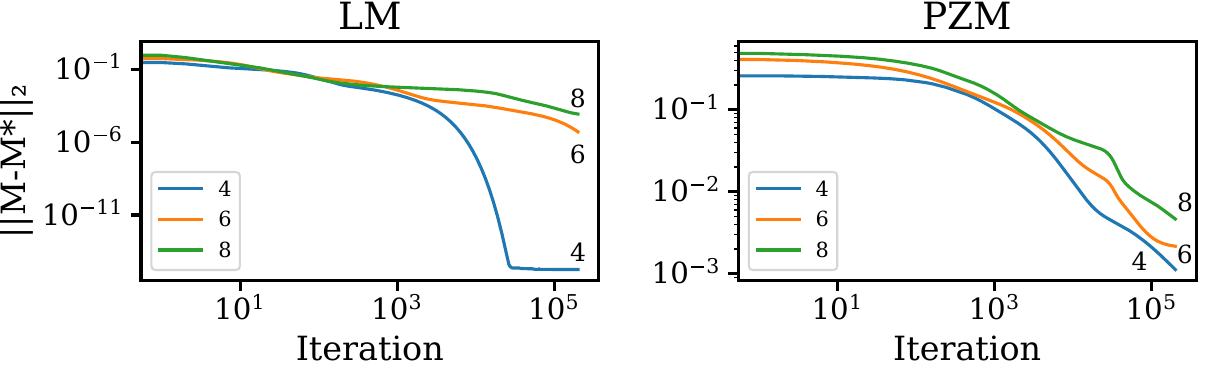}
\caption{The 2-norm of the moment vector error for varying maximum orders of LMs and PZMs using the controller with perfect moment estimates.}
\label{fig:MSREConErr}
\end{figure}

The gain matrix $\Gamma$ was determined experimentally, with the observation that higher-order moments should have smaller associated gains for faster convergence. We define 
$\Gamma=G$ as
\begin{equation}
G = \operatorname{diag}(g) \in \real^{\mathfrak{m}\times \mathfrak{m}},     \label{eq:controlgain}
\end{equation}
\[
    g = (\underbrace{\{{1^{-1.7}},{1^{-1.7}} \}}_{d=1}, \underbrace{\{{2^{-1.7}},{2^{-1.7}},{2^{-1.7}}\}}_{d=2}, \ldots,\underbrace{\{{d^{-1.7}}\}_{2n-1}}_{d=n}),
\]
i.e., $d+1$ copies of $d^{-1.7}$ for $d=1,\ldots,n$, the maximum degree in the moment vector. The exponent $-1.7$ was found experimentally and is a tunable parameter chosen to create smaller control gains for higher-order moments.

LMs and PZMs up to orders $2$, $4$, $6$, and $8$ were calculated for the ``bunny head'' distribution depicted in Figure~\ref{fig:experiment_shapes} and were used as the goal moment vectors. Each control trial consisted of 200,000 iterations from the same initial robot configuration. Figure~\ref{fig:MSREConErr} shows the robots' convergence to the desired distribution.

\subsection{Simultaneous Estimation and Control} 
Even if each robot's estimator is guaranteed to converge to the formation's actual moments and the controller is guaranteed to converge to the desired moments using perfect estimates of the swarm moments, when we couple estimation and control the system may lose these properties and exhibit more complex behaviors. As an example, for $N=7$ robots ($2N=14$) and third-order LMs ($\mathfrak{m} = 9$ moments), we simulated: the estimator alone (stationary robots) with estimator gain $\gamma = 1/(10N)$, memory state, and no packet loss; the controller alone with perfect moment estimates and controller gains $\Gamma_1 = 5 G$ and $\Gamma_2 = G$, where $G$ is given by Equation~\eqref{eq:controlgain}; and combined estimation and control with $\Gamma_1 = 5 G$ and $\Gamma_2 = G$. 

\begin{figure} 
\centering
\includegraphics[width=0.9\columnwidth]{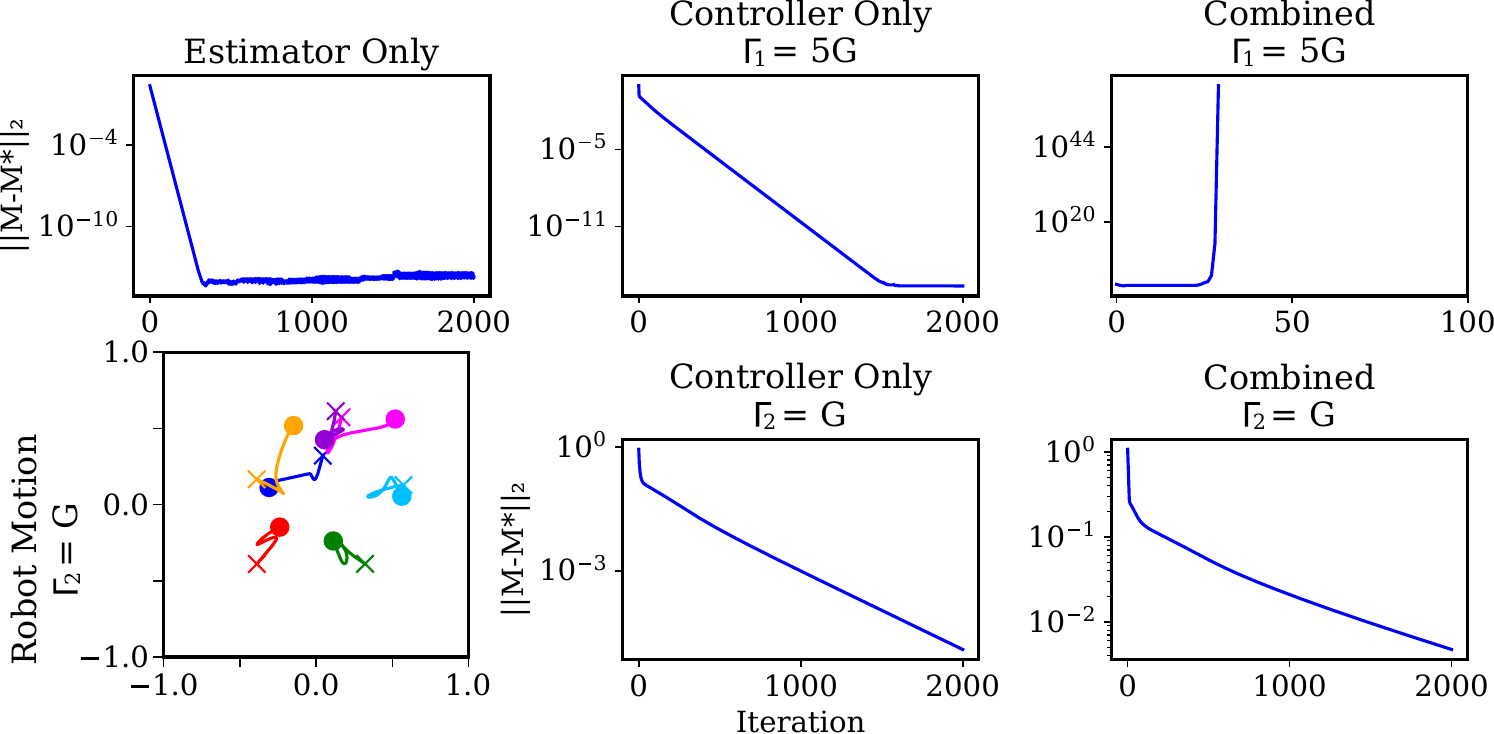}
\caption{Comparing estimation alone, control alone, and combined estimation and control for two different control gain matrices. Top left: The 2-norm of the moment vector error converges to zero for one robot's estimates vs.\ desired moments. Bottom left: The motions of the robots (from x's to o's) for control gains $\Gamma_2 = G$ and perfect moment estimates. Middle column: Performance of the controller with perfect moment estimates. The robots converge to the desired moments for both control gains $\Gamma_1 = 5G$ and $\Gamma_2 = G$. Right column: Performance of the controller when each robot uses estimated moments. The robots converge to the desired moments for the smaller control gains but not the larger gains.}
\label{fig:ControlGainEffect}
\end{figure}

Figure~\ref{fig:ControlGainEffect} plots the results of 2000 iterations. The estimator converges to the correct moment estimates with stationary robots. The controller with perfect estimates drives the robots to the desired moments for control gains $\Gamma_1$ and $\Gamma_2$. The combined estimation and control system is unstable for control gains $\Gamma_1$ but convergent for $\Gamma_2$. This is an example of aggressive control gains coupled with estimator delay causing instability. Future work may include a stability analysis, e.g., using small-gain theory or passivity analysis, to better understand how aggressively to choose motion control gains in light of delays inherent in distributed estimation.

\section{Experimental Implementation} 

\subsection{Coachbot~V2.0 System}
Experiments were performed with the Coachbot~V2.0 swarm setup, which consists of 100~differential drive robots and a high-fidelity Docker simulation for testing robot control algorithms before deployment~\cite{Wang2020}. The robots sense their positions and orientations using time-varying infrared signals from a ceiling mounted HTC VIVE lighthouse and communicate with each other and a centralized workstation via Wi-Fi with an optional artificially-imposed limited communication radius. In practice, we found that 40-50\% of communicated packets are dropped due to the high density of communication.

\begin{figure}
\centering
\begin{subfigure}{\linewidth}
    \centering
    \includegraphics[height=.21\linewidth]{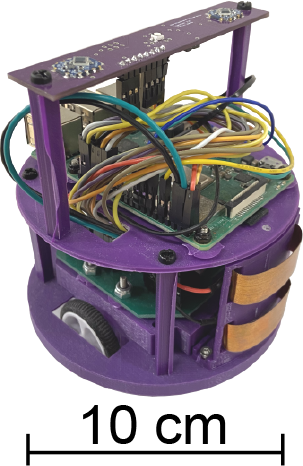}
    \hspace{0.5cm}
    \fbox{\includegraphics[width=.18\linewidth]{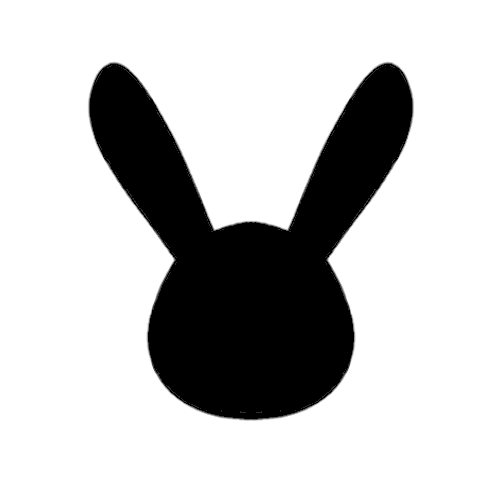}}
    \hspace{0.5cm}
    \fbox{\includegraphics[width=.18\linewidth]{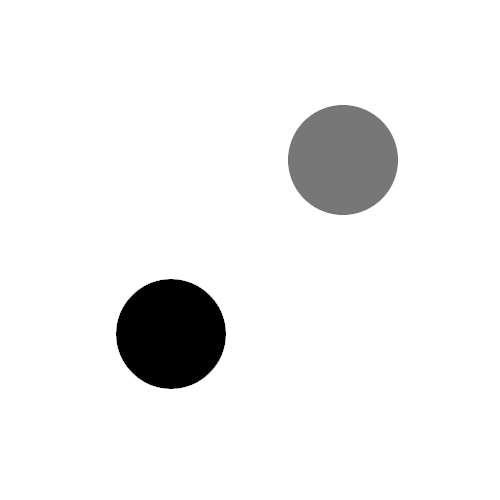}}
    \caption{Left: Coachbot robot. Middle: Uniform-density ``bunny head'' formation. Right: Formation described by two identically-sized disks. The darker disk is twice the density of the lighter disk.}
    \label{fig:experiment_shapes}
\end{subfigure}
\par\bigskip
\begin{subfigure}{\linewidth}
    \centering
    \includegraphics[width=0.9\linewidth]{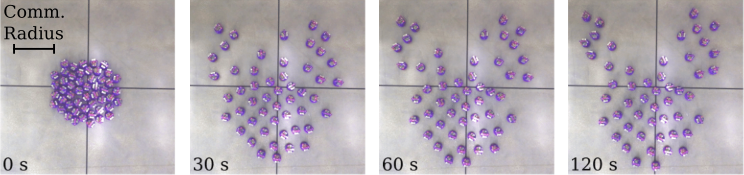}
    \caption{Fifty Coachbots forming the bunny head using eighth-order LMs.}
    \label{fig:leBunnyProg}
\end{subfigure}
\par\bigskip
\begin{subfigure}{\linewidth}
    \centering
    \includegraphics[width=0.9\linewidth]{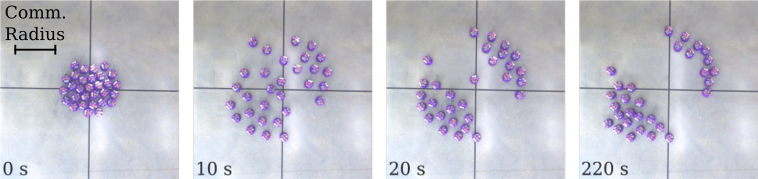}
    \caption{Thirty Coachbot robots forming the disks using sixth-order PZMs.}
    \label{fig:pzmGCProg}
\end{subfigure}
\caption{Experimental desired formations.}
\end{figure}

Every 0.27~s, each robot executes the control loop of Figure~\ref{fig:block_diagram}: it senses its position, updates the estimator input~\eqref{eq:estInput}, checks for received messages, computes the estimator~\eqref{eq:estimator}-\eqref{eq:estimator2}, broadcasts estimator signals, and calculates the control velocity of a reference point on the robot away from the wheel axis~\eqref{eq:controllaw}. This velocity is modified by a collision-avoidance filter (based on three zones of varying repulsion around each robot that prevents collisions with neighbors~\cite{Hwang2013}) and then bounded. Velocities below a minimum threshold are set to zero. The calculated velocity of the robot reference point is transformed to wheel speeds as described in~\cite{Hwang2013,ModernRobotics}.

The experimental system differs from the model in that (1) there is no global clock and (2) the robots have nonzero extent and implement collision avoidance and velocity saturation. 

Below we report experimental distributed formation control with two goal formations: a binary ``bunny head'' formation, where the desired density inside the bunny head silhouette is constant and zero outside, and a bimodal formation consisting of two identically-sized solid disks, with the density of one disk twice the other's (Figure~\ref{fig:experiment_shapes}). The arena is 3~m $\times$ 3~m in size, for a scale factor of 1.5~m relative to the $[-1,1]^2$ LM and PZM representations. The communication radius was set to 1.5~m. In all experiments, the robots started in a random configuration centered around the arena origin. Multiple trials were conducted for each experiment. The results shown in this paper are representative of typical results. Due to limited space, only the bunny head with LMs and the two-disks with PZMs are reported; other experiments can be seen in the video~\cite{video}.

Controller gains were $\Gamma = G$ from~\eqref{eq:controlgain}. The estimator gain $\gamma$ was chosen to be $1/N$. Memory kept neighbor messages for up to $1.5N$ consecutive iterations.  Desired moments were calculated from the images of the bunny head and two-disk formations.   
The following results show the moment estimates of robot 15 (simulated or real), which are representative of each robot, as each robot's moment estimates converged to the actual moments.

\subsection{Binary ``Bunny Head'' Formation}

\begin{figure}  
\centering
\includegraphics[width=\linewidth]{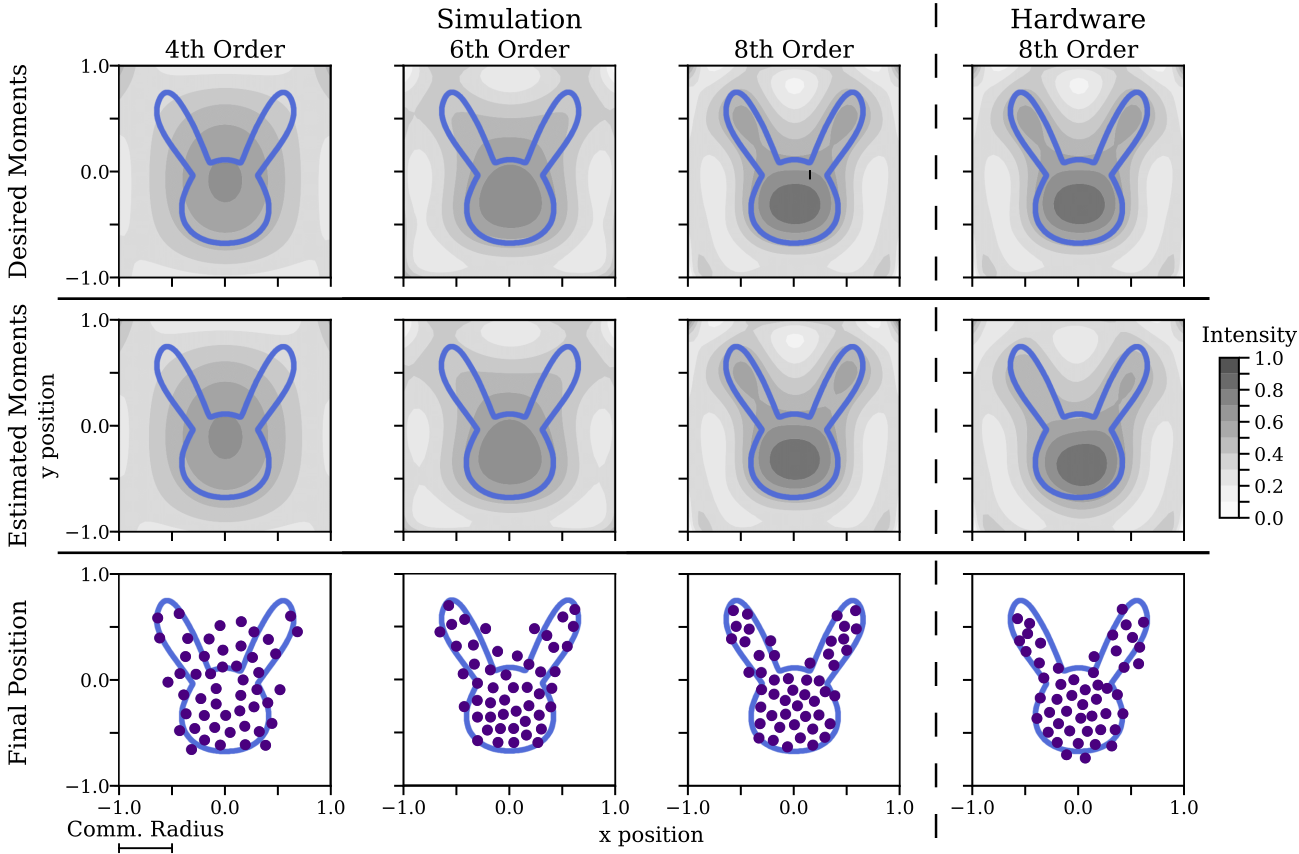}
\caption{Top row: Grayscale image reconstructions using the fourth, sixth, and eighth-order LM representations of the bunny head formation. Middle row: Grayscale image reconstructions of the final estimated moments after distributed estimation and control by 50 robots in simulation (fourth-, sixth-, and eighth-order moments) and experiment (eighth-order moments). Bottom row: The final robot configurations.}
\label{fig:lebunny}
\end{figure}

\begin{figure} 
\centering
\includegraphics[width=0.85\columnwidth]{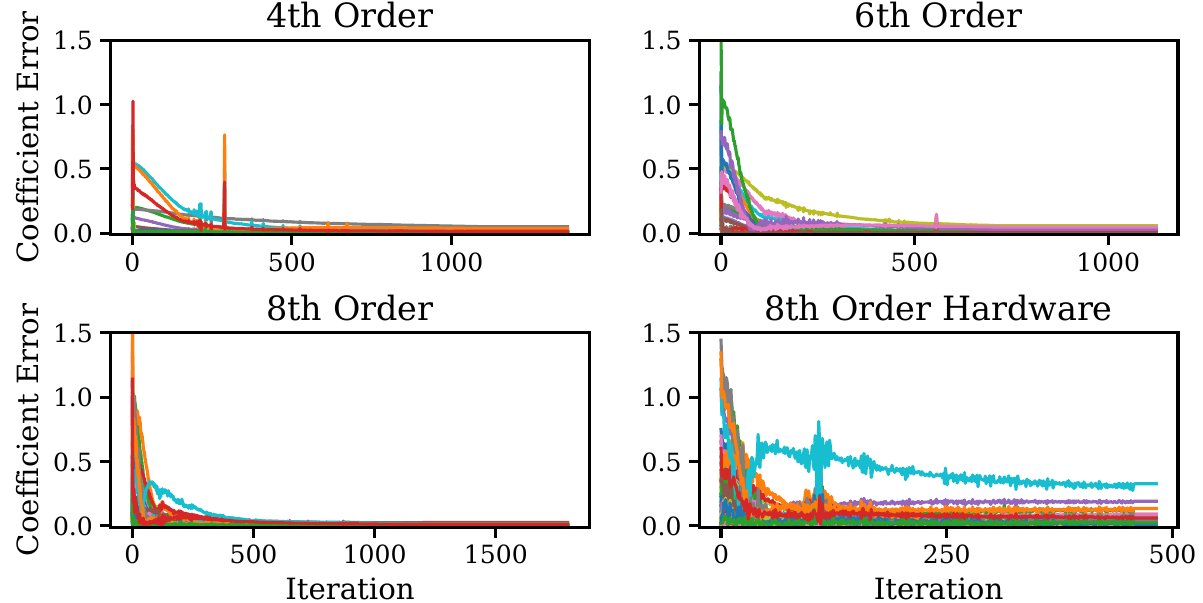}
\caption{Absolute value of the difference between robot 15's moment coefficient estimates and the desired moments for increasing orders of LMs representing the bunny head shape.}
\label{fig:lebunnyCoeffErr}
\end{figure}

\begin{figure} 
\centering
\includegraphics[width=0.8\columnwidth]{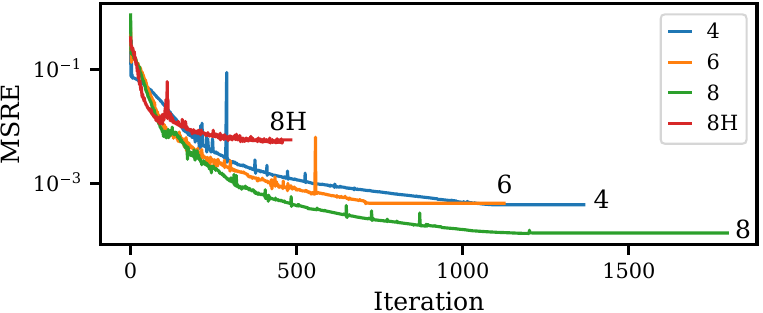}
\caption{MSRE vs.\ control iterations for the bunny head using LMs of increasing maximum order (fourth-, sixth-, and eighth-order moments) and experiment 8H (eighth-order moments).}
\label{fig:lebunnyErrLog}
\end{figure}

\begin{figure} 
\centering
\includegraphics[width=0.70\linewidth]{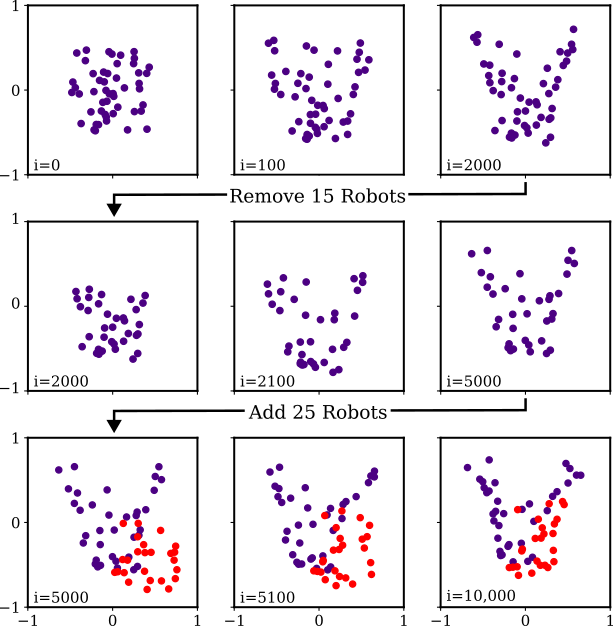}
\caption{Top row: 50 simulated robots form the bunny head after 2000 iterations using sixth-order LMs. Middle row: After instantly removing 15 robots, the swarm self-heals, regrowing the bunny ears at 5000 iterations. Bottom row: After instantly adding 25 robots (shown in red), the swarm adjusts to form the bunny head at 10,000 iterations. }
\label{fig:subAddRob}
\end{figure}

Figure~\ref{fig:leBunnyProg} shows snapshots of 50~Coachbots moving toward the eighth-order LMs of the bunny head formation. Figure~\ref{fig:lebunny} shows the distributions reconstructed from LMs up to fourth-order (14 moments), sixth-order (27 moments), and eighth-order (44 moments), respectively, as grayscale images; the final simulated configurations of the robots using the estimator~\eqref{eq:estimator}-\eqref{eq:estimator2} and control law~\eqref{eq:controllaw}, as well as the corresponding grayscale representations of the reconstructions from the LMs; and the final robot configurations and the 
reconstructed distribution for a hardware experiment with eighth-order LMs.

The fidelity of the moment representation of the bunny head increases as the moment order increases. The experiments also show that distributed estimation and control is effective in driving the robots to achieve the desired moments.  
The individual LM coefficient errors are shown in Figure~\ref{fig:lebunnyCoeffErr}. 
The larger error in the eighth-order hardware relative to the simulation is primarily attributed to the velocity deadband in experiment.

Figure~\ref{fig:lebunnyErrLog} shows the MSRE decreasing as the robots converge. 
As the error decreases, the commanded velocities reach the deadband, preventing further decrease of the MSRE.

To demonstrate the impact of self-healing moment estimation on formation control, we simulate 50 robots forming the bunny head using sixth-order LMs for 2000 iterations, then instantly remove 15 robots and continue the simulation (Figure~\ref{fig:subAddRob}). The swarm regrows the bunny ears. After 5000 iterations, we instantly add 25 robots, shown in red, to the bottom right quadrant. The swarm adjusts to form the bunny head with the new robots.

\subsection{Two-Disk Formation}
\begin{figure}
\centering
\includegraphics[width=0.95\linewidth]{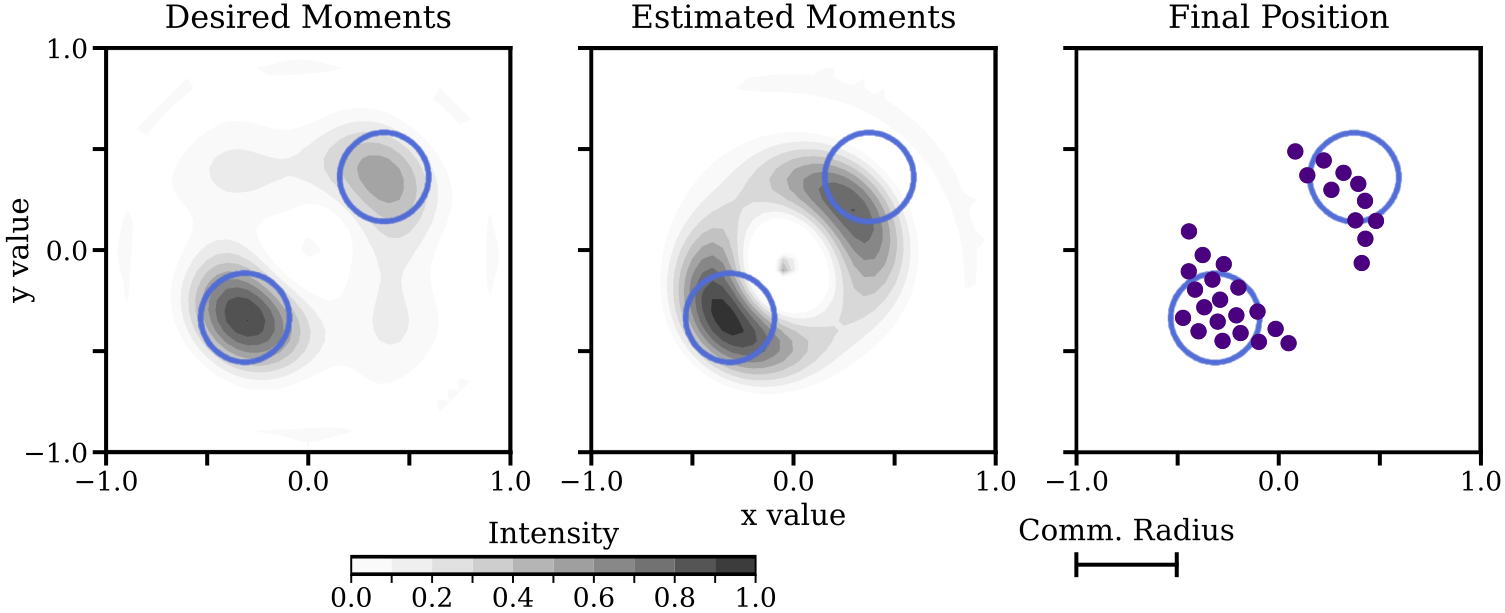}
\caption{Two-disk formation represented by sixth-order PZMs and the final experimental configuration of 30~robots.}
\label{fig:pzm_greycirc}
\end{figure}

\begin{figure} 
\centering
\includegraphics[width=0.8\columnwidth]{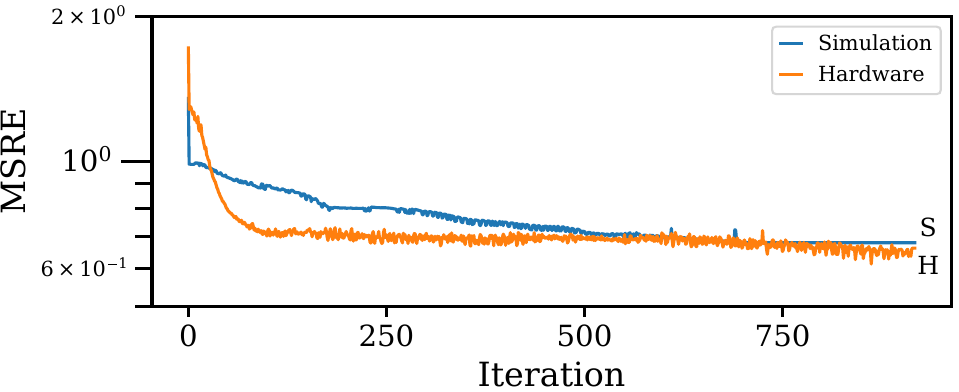}
\caption{MSRE of the two-disks using sixth-order PZMs.}
\label{fig:pzm_greycircErr}
\end{figure}

Figure~\ref{fig:pzmGCProg} shows snapshots of 30~Coachbots moving toward the sixth-order PZMs of the two-disk formation, and Figure~\ref{fig:pzm_greycirc} shows the reconstruction of the desired moments, the reconstruction of the estimated moments, and the final configuration achieved by the 30~robots. The robots clearly segregate to the two regions, with a higher number of robots (19 vs.\ 11) in the region with twice the desired density. In the experiment, the size of the robots, coupled with their collision avoidance, actually prevents 20~robots from fitting inside the boundaries of the denser disk. Figure~\ref{fig:pzm_greycircErr} shows the robots converged to their final positions with a small steady-state MSRE. 

\section{Conclusion}
\label{sec:conclusion} 

This paper presents a method for distributed, scalable, self-healing swarm formation estimation and control based on image moment representations, where the level of detail in the desired formation representation is determined by the number of image moments. In simulation and experiments with up to $50$ robots, the robots approximately achieve specified formations despite nearly $50$\% packet loss.

Further work is needed to determine how to choose estimator and controller gains that ensure good coupled performance. The gains used in this paper were found empirically. Generally, using smaller control gains on higher-order moments, which encode high-frequency features, results in better convergence.
Beyond gradient control, distributed  model predictive control could improve the rate of convergence to desired moments.

\bibliographystyle{IEEEtran}
\balance
\small
\Urlmuskip=0mu plus 1mu\relax
\bibliography{moments.bib, formulation.bib, references.bib}

\end{document}